# Learning Hidden Patterns from Patient Multivariate Time Series Data Using Convolutional Neural Networks: A Case Study of Healthcare Cost Prediction

Mohammad Amin Morid, PhD[1], Olivia R. Liu Sheng, PhD[2], Kensaku Kawamoto, MD, PhD, MHS [3], Samir Abdelrahman, MS, PhD[3,4]

[1] Leavey School of Business, Santa Clara University, Santa Clara, CA, USA;

[2] David Eccles School of Business, University of Utah, Salt Lake City, UT, USA;

[3] Department of Biomedical Informatics, University of Utah, Salt Lake City, UT, USA;

[4] Computer Science Department, Cairo University, Giza, Egypt.

## Abstract

*Objective*: To develop an effective and scalable individual-level patient cost prediction method by automatically learning hidden temporal patterns from multivariate time series data in patient insurance claims using a convolutional neural network (CNN) architecture.

*Methods:* We used three years of medical and pharmacy claims data from 2013 to 2016 from a healthcare insurer, where data from the first two years were used to build the model to predict costs in the third year. The data consisted of the multivariate time series of cost, visit and medical features that were shaped as images of patients' health status (i.e., matrices with time windows on one dimension and the medical, visit and cost features on the other dimension). Patients' multivariate time series images were given to a CNN method with a proposed architecture. After hyper-parameter tuning, the proposed architecture consisted of three building blocks of convolution and pooling layers with an LReLU activation function and a customized kernel size at each layer for healthcare data. The proposed CNN learned temporal patterns became inputs to a fully connected layer. We benchmarked the proposed method against three other methods: 1) a spike temporal pattern detection method, as the most accurate method for healthcare cost prediction described to date in the literature; 2) a symbolic temporal pattern detection method, as the most common approach for leveraging healthcare temporal data; and 3) the most commonly used CNN architectures for image pattern detection (i.e., AlexNet, VGGNet and ResNet) (via transfer learning). Moreover, we assessed the contribution of each type of data (i.e., cost, visit and medical). Finally, we externally validated the proposed method against a separate cohort of patients. All prediction performances were measured in terms of mean absolute percentage error (MAPE).

*Results:* The proposed CNN configuration outperformed the spike temporal pattern detection and symbolic temporal pattern detection methods with a MAPE of 1.67 versus 2.02 and 3.66, respectively ($p<0.01$). The proposed CNN outperformed ResNet, AlexNet and VGGNet with MAPEs of 4.59, 4.85 and 5.06, respectively ($p<0.01$). Removing medical, visit and cost features resulted in MAPEs of 1.98, 1.91 and 2.04, respectively ($p<0.01$).

*Conclusions:* Feature learning through the proposed CNN configuration significantly improved individual-level healthcare cost prediction. The proposed CNN was able to outperform temporal pattern detection methods that look for a pre-defined set of pattern shapes, since it is capable of extracting a variable number of patterns with various shapes. Temporal patterns learned from medical, visit and cost data made significant contributions to the prediction performance. Hyper-parameter tuning showed that considering three-month data patterns has the highest prediction accuracy. Our results showed that patients' images extracted from multivariate time series data are different from regular images, and hence require unique designs of CNN architectures. The proposed method for converting multivariate time series data of patients into images and tuning them for convolutional learning could be applied in many other healthcare applications with multivariate time series data.

## 1. INTRODUCTION

Rising U.S. healthcare spending has been become a serious issue for individual, institutional and government payers and financiers of healthcare [1,2]. Government agencies such as the Centers for Medicare & Medicaid Services (CMS) and healthcare organizations are actively engaged in quality improvement and cost-saving efforts to help contain increasing healthcare costs while maintaining care quality. These healthcare stakeholders have a keen interest in accurately measuring the impact of quality and cost improvement solutions. Accurate healthcare cost prediction can provide an essential foundation for directing care management resources and evaluating efforts at reducing healthcare costs [3]. Furthermore, patient cost prediction could benefit healthcare financial planning undertaken by patients, their families, healthcare delivery systems, and payors. A scalable, accurate cost prediction method could be a valuable addition to the cost estimation tools provided by insurance companies and healthcare providers to their customers [4].

The problem of individual healthcare cost prediction has drawn growing research attention from the medical informatics and healthcare management research communities [5]. Recent studies [6,7] have suggested that supervised learning methods applied to insurance claims data can be used to predict future costs. However, these prior studies still are not optimal, with even the best models. To improve the accuracy of cost prediction over extant solutions [7], we previously leveraged multivariate fine-grain time series of cost and non-cost (e.g., diagnosis and visit) data to create advanced temporal features by detecting non-linear trends (e.g., spikes) in each time series for supervised predictive model learning for healthcare costs. This approach, however, is difficult to scale due to the need to manually craft model features. To overcome this limitation, this study focused on exploring the potential to leverage deep neural network architectures for feature learning to improve the scalability of patient cost prediction while increasing accuracy.

The deep learning literature [8–10] has identified the potential of the convolutional neural network (CNN) architecture to represent learning from grid-like data, including image, speech and time series data. Prior studies that applied the CNN architecture to multivariate time series data used the initial network layers primarily to learn temporal patterns from univariate time series data. The integration of multiple univariate feature maps occurred at a late stage in the CNN architectures of those studies [11–13]. In many imaging data mining applications, CNN architectures have been able to effectively learn latent (i.e., hidden) two-dimensional (2D) patterns from 2D image data that improve the performance of image classification models [14,15]. The ability of the CNN architectures to learn regional spatial features throughout the CNN layers could be beneficial to representation learning from other varieties of 2D data. In particular, by representing multivariate time series data in a 2D matrix format of data along the variable and time dimensions, a CNN could potentially learn hidden abstract pattern features from the relationships underlying multiple variables over multiple time segments. To the best of our knowledge, the extant deep learning and medical informatics literature has not explored the predictive power of latent 2D patterns in multivariate time series data in such a manner. Because the pre-specified features for state-of-the-art cost prediction models [3,7] include lines, triangles and other geometric patterns in a single time series, and CNNs are suited for 2D or three-dimensional (3D) spatial pattern learning, the focus of this study on proposing an effective CNN configuration to learn such patterns, among other latent patterns, from multivariate time series patient data is fitting and also a research gap.

One of the main challenges in learning latent 2D features from the time series of multiple variables is that the order or positions of the variables along the variable dimension of an input matrix should not have the same meaningful impact as that of a regular 2D image matrix on latent 2D feature learning. Hence, this study explored the effects of various hyper-parameters of a conventional CNN architecture to identify an effective CNN configuration that significantly improved the performance of patient cost prediction using insurance claims data over the closest benchmarks, which employed manually-crafted advanced time series patterns. Because of the misuse of variable order or positions in a multivariate time series input matrix in feature learning, the competition-winning CNNs – AlexNet [35], VGGNet [14] and ResNet [36] – could not produce more accurate patient cost prediction than the proposed CNN and the prior benchmarks. Rather, by setting the variable dimension of the convolution kernel filters in a conventional CNN architecture to the total number of input variables, the training of the proposed CNN configuration relaxed the constraints of learning latent 2D features from the time series of multiple variables based on their positions or order.

Our findings contribute to health informatics research and practice by describing a more accurate and scalable patient cost prediction approach compared to prior machine learning approaches for this problem. Moreover, as the first study that has shown the benefits of learning latent 2D time series patterns for patient cost prediction, many future studies can build on our findings to advance the CNN architectures or other representation learning approaches to further enhance the accuracy and scalability of deep-learning-based health analytics.

## 2. BACKGROUND

### 2.1 Cost prediction

Our literature review [6] showed that prior research on the use of supervised learning for cost prediction fell into three types. The research goal of the first type of cost prediction studies was to predict cost using patients' clinical features, such as chronic disease scores, and to examine their effects on cost prediction [16]. The second type of cost prediction research examined the predictive power of cost predictors with or without non-cost predictors (e.g., medical or visit data) extracted from electronic medical records or medical insurance claims. In the last type of cost prediction research literature, researchers bucketed individuals' medical costs and developed models that predict the cost bucket to which a patient's cost should belong. Medical cost bucket prediction could guide various decisions, such as the allocation of care coordination resources, based on a patient's predicted future costs.

This study extended the approaches in the second type of individual-level patient cost prediction using cost predictors and other non-cost and non-clinical predictors using medical insurance data. The literature review in [6] identified five prior research studies [1,3,5,17,18] that used this range of features to predict future patient healthcare costs. Morid et al. [7] extracted multivariate time series features from patients' cost, diagnosis code, prescription and visit records. These features along with advanced manually-crafted features based on change patterns (i.e., spike patterns over time), lifted the prediction performance over all other studies they identified. We benchmarked the prediction performance enabled by the proposed CNN against the model developed in this prior study [7].

## 2.2 Temporal pattern detection

Healthcare claims data contain rich time series data, thereby providing an important opportunity to discover new knowledge using various data mining methods. However, common multivariate time series prediction methods cannot be directly applied to such healthcare temporal data for several reasons [19]. Data points in the time series of claims data are often sampled or recorded at different frequencies for different patients. Also, large amounts of missing data points are common due to intentional causes (e.g., medical reasons) or unintentional causes (e.g., human mistakes) [20].

The most common approach to overcome these issues is to transform the raw data of the multivariate time series into a standard form in which the time series data are uniformly represented [21] using two types of transformation: static and dynamic [20]. Using the static transformation (ST) approach, each time series is represented by a pre-defined set of features and their values (e.g., most recent platelet measurement, maximum hemoglobin measurement). After this transformation, numerical prediction methods can be applied to predict the desired outcome. Although the ST approach facilitates the prediction process by reducing dimensionality, it also results in information loss by ignoring the temporal trends in the time series, which could affect the prediction accuracy [22]. An alternative approach, dynamic transformation (DT), segments each time series into equal-sized windows and represents each window by a statistical measure of the events in the window. This approach can then extract a pre-defined set of temporal pattern shapes over windows of equal size from the transformed multivariate time series [23].

The most common DT approach for pattern detection is to extract symbolic patterns from symbolic multivariate time series. These patterns are generally referred to as time-interval-related patterns (TIRPs) in the literature [24–26]. The most common approach to extract TIRPs is using Allen's temporal relations [27], with seven relations to capture the state of two alphabetic time intervals against each other (e.g., *overlap*, *equals*, *meets*). Several studies have attempted to use all or part of these relations for pattern extraction [20,28–30]. Moskovitch and Shahar [19] proposed a fast time-interval mining method, called KarmaLego, to exploit temporal relations [19]. KarmaLego consists of two main steps: Karma and Lego. In the Karma step, all of the frequent two-sized TIRPs are discovered using a breadth-first-search approach. In the Lego step, the frequent two-sized TIRPs are extended into a tree of longer, frequent TIRPs. Recently, the same authors proposed a set of three abstract temporal relations (i.e., *before*, *overlap* and *contain* as disjunctions of Allen's relations) and showed that it is more effective than using the full set of Allen's relations [31]. They called their general framework for classification of multivariate time series analysis KarmaLegoSification (KLS). In this paper, we used KLS with the three temporal relations as a baseline to implement symbolic pattern detection. Symbolic pattern detection has been shown to be successful in a variety of healthcare applications, such as predicting patients who are at risk of developing heparin induced thrombocytopenia [32], hospital length of stay predictions using health insurance claims data [33] and prediction of stroke [34].

To the best of our knowledge, the most accurate research (i.e., most relevant study) to date on cost prediction has been that by Morid et al. [7] (the authors of this paper), who propose a temporal pattern extraction. This prior work used change point pattern detection to recognize spikes in patients' cost profiles and applied a gradient boosting classifier to predict patients' future costs. The idea behind this method was to distinguish the constant high cost pattern of chronic patients from the temporary high cost pattern of patients with exceptional situations (e.g., pregnancy or accident). The assumption of this approach was that the latter cost pattern that exhibits a spike might have a low risk of high future costs. This method is the second baseline used in this paper.

Although the DT approach reduces information loss and increases accuracy, its performance is dependent on a pre-defined set of temporal pattern shapes, which may not be easy to define in all healthcare contexts [7]. For instance, the method described by Moskovitch and Shahar [19] is limited to Allen's relations (i.e., before, overlap, contain) and the method described by Morid et al. [7] is limited to change point detection patterns (i.e., spike pattern shapes). In other words, all these studies make assumptions about the specific type (i.e., shape) of patterns researchers look for based on their healthcare domain knowledge. In this study, we proposed a CNN method that could learn and leverage hidden patterns for patient cost prediction without any assumption or prior domain knowledge to manually engineer features.

**2.3 Convolutional neural network**

The demonstrated successes of CNNs for learning 2D spatial features of image data [35] motivated the objective and methods of this study. The layers of a CNN architecture are trained to gradually detect more and more complex patterns via three architectural ideas: local receptive fields, shared weights and spatial subsampling [7]. For instance, for human image classification, each image consists of components like eye, mouth, nose, etc. Each component is constructed from various kinds of lines, such as straight and curved. CNNs first try to identify those lines, which represent some sort of spatial pattern, build the components (e.g., eye or mouth) on top of them, and finally make the prediction. This way, the input for the final prediction is a meaningful set of patterns rather than raw pixel values. Similarly, this idea can be employed in the healthcare domain to represent the multivariate time series of healthcare data (e.g., patients' costs in the past, vital signs, lab tests) by their temporal patterns, followed by use for outcome (e.g., cost) prediction [36].

A CNN typically consists of an input layer, one or more blocks of convolution layer(s) and pooling layer(s) and a fully connected layer. The input layer retrieves the raw data (e.g., image pixels) as input to a convolution layer. Using convolution kernel filters to map each 2D sub-region of the input data into a scalar output, the convolution layer produces a higher-level feature representation (i.e., feature map) of the input data. To discretize the values, the feature map passes through a linear activation function. To a varying extent, the convolutional layer functions as the feature extraction stage of traditional pattern detection methods (e.g., spike pattern detection). The main difference is that the model discovers the patterns without using pre-defined pattern shapes such as spikes, as discussed in prior work [37]. To reduce the data dimension (i.e., feature reduction), the pooling layer performs a down sampling operation by summarizing data in different regions. Finally, the fully-connected layer is similar to a fully connected neural network that generates the final output. More details about CNNs can be found in [38].

Some CNN architectures, including AlexNet [35], VGG-16 [15] and ResNet-34 [39], have been shown to successfully detect patterns in images. AlexNet consists of five convolutional layers and three fully connected layers with kernels of 11×11, 5×5 and 3×3 in dimensions, max pooling, dropout and ReLU activation functions. This CNN architecture won the ImageNet competition in 2012. VGG-16 is a CNN architecture with 13 convolutional layers, all with 3×3 kernel sizes and combinations of 64, 128, 256 and 512 filters augmented with two fully connected layers and with 4096 nodes at the end. With the idea that deeper networks are capable of learning more complex patterns that should lead to better performance, ResNet-34 consists of 34 convolutional layers with the same kernel size and filter sizes as VGG-16. Although these architectures have been proposed for a specific application (e.g., ImageNet competition), they can be used for other similar applications. With transfer learning (TL), which is commonly used for this purpose, a deep learning architecture created for one problem is applied to a similar problem [40]. Since deep learning can automatically learn the features and preserve the knowledge of the best patterns on the network layers and parameters, TL can reuse the trained model and transfer the patterns.

Diagnosis of pediatric pneumonia with chest X-ray images using ResNet [41], interstitial lung disease detection with computerized tomography (CT) scan images using AlexNet [42], and age-related macular degeneration (AMD) severity classification with fundus images using VGGNet [43] are some examples of applications for which TL has been effective. Also, a variety of non-image problems in healthcare such as diabetes detection from heart rate signals [44] and Parkinson's disease detection using voice recordings [45] have been addressed via models that utilize TL approaches to leverage some of the pre-trained models using ImageNet data. To assess the potential and limitations of these image classification CNN architectures for learning latent 2D multivariate temporal patterns, this study conducted an experiment that applied AlexNet [35], VGG-16 [15] and ResNet-34 [39] based on transfer learning to our application (i.e., patients' multivariate time series matrices). Employing well-trained CNN architectures could have a significant advantage over proposing a new CNN, since these architectures do not need training from scratch and can be used in clinical settings with relatively small amounts of data.

Utilizing CNN architectures for representation learning from multivariate time series data has received limited attention in video, motion or healthcare deep learning research. [11] was the first study to define and apply a one-sided convolution operation over the events or periods along the time axis of a single time series in the input matrix of multivariate time series data from electronic health records (EHRs) to learn one-dimensional (1D) temporal patterns. Subsequently, [36] followed and applied such one-sided convolution filters to EHR data for latent temporal pattern discovery, which improved the prediction of patients' risks of congestive heart failure and chronic obstructive pulmonary disease over the baselines. Working with multivariate time series of electroencephalographic (EEG) recordings, [12,13] also employed one-sided convolution kernel filters to extract 1D temporal patterns from different data sets. Applications of one-sided filters to extract temporal patterns along the time dimension can also be found in prior literature on action and mobility detections in video and sensor data [46–48]. Even though these studies proposed different algorithmic innovations in the CNN architectures, their methods limited, to some extent, the learning of latent 2D patterns over the time and multivariate dimensions simultaneously. This study proposed the use of two-sided convolution filters along the variable and time dimensions in CNN architectures to leverage joint temporal patterns across different time series.

## 3. METHODS

To learn latent 2D temporal patterns from patient data, the multivariate time series data of each patient is shaped into the form of an image (i.e., a matrix with time windows on one dimension and the medical, visit and cost features on the other dimension). To collect the input features for the patients, aligned with recent studies on cost prediction [7], we used medical, visit and cost data, aligned with a recent study on cost prediction [7]. More specifically, we used an input matrix of patient cost, including the time series of the 180 procedure groups and the 336 drug groups used by Bertsimas et al. (2008) [3], the 83 diagnosis groups used by Duncan et al. (2016) [1], the seven visit categories used by Morid et al. (2019) [7] and two categories of costs (medical and pharmacy costs). Table 1 summarizes these features.

Table 1 - Time series features used as inputs for this study

| Category | Time Series Features | Cost Inputs | Reference |
|---|---|---|---|
| Cost | 2 | Medical cost and pharmacy cost | |
| Medical | 180 | Procedure groups | Bertsimas et al. (2008)[3] |
| Medical | 83 | Diagnosis groups | Duncan et al. (2016) [1] |
| Medical | 336 | Drug groups | Bertsimas et al. (2008)[3] |
| Visit | 7 | Office, Inpatient, Outpatient, Lab, Emergency, Home, Other | Morid et al. (2019) [7] |

Aligned with previous studies on temporal pattern detection [7,49], each time series was divided into fixed size windows. To prepare a patient's input matrix, the total amount of patient costs in each cost group (e.g., medical and pharmacy), the total number of patient visits in each visit group (e.g., inpatient and outpatient) and the total number of the patient claims in each medical group (e.g., procedure, diagnosis and drug) were computed to represent the values in the cells belonging to a time window. After evaluating different window sizes ranging from 2 weeks to 6 months, the 1-month window size based on its performance. Since two years of patients' data were used to predict their cost in the following year, a patient's input matrix included 608 x 24 data values.

For model training, the input matrix of multivariate time series patient data was feed-forwarded through various convolution, detector (non-linear activation) and pooling layers, followed by back-propagation to update weights in the network. Aligned with previous studies, one sequential combination of convolution and pooling layers is referred to as a *block* in this study [38]. The feed-forward process in a convolution layer applied a kernel, which is a rectangle matrix of $a \times b$ elements, to each $a \times b$ sub-region of cells in the input matrix and output the convolutional result: the dot matrix product of the kernel and each of the sub-regions of $a \times b$ input elements. Here, $a$ and $b$ are typically smaller than the numbers of rows and columns of an input matrix. In a supervised learning task such as our patient cost prediction model training, the elements of the kernel were learned from the training data. An important design decision in our proposed CNN is that the kernel dimensions are set to be $f \times k$, where $f$ is the total number of time series variables or features in the input matrix to a convolution layer, and $k$ (a value between 1 to 9) is the length (i.e., the number of time windows) of the target temporal patterns. The underlying idea is that the kernel can convolve over all input variables across different time windows to find 2D temporal patterns of increasing predictive power over various CNN layers. For instance, a CNN with $k=1$ detects monthly temporal patterns in the first convolutional layer, and similarly a CNN with $k=3$ begins with detecting temporal patterns in each three-month period. More specifically, assuming each patient's claims data are represented by an $f \times t$ matrix, $x_{1:f,i:j}$ is a sub-region of the input matrix corresponding to the values of all $f$ features in time window $i$ to time window $j$. Here, $j - i = k$. Let $t$ denote the total number of time windows per time series. The first convolution layer convolves a 2D kernel of $f \times k$ values over each sub-matrix $x_{1:f,i:j}$, where $i = 1...t-k$, to produce an output matrix. Each value $c_d$ in the output matrix is generated as $c_d = F\_{activation}$ (w. $x_{1:f,i:j}$ + b), where w. $x_{1:f,i:j}$ is the dot-product of $x_{1:f,i:j\ j}$ with the kernel weights in $w$, $b$ is a bias term and $F\_{activation}$ is an activation function.

While the sigmoid function is the most commonly used activation function for neural networks, including CNNs, it was recently shown that the saturation problem makes the function perform poorly for training neural network models [50]. To address this limitation, the rectified linear unit (ReLU) activation function can be used to accelerate the convergence of stochastic gradient descent. However, ReLUs also suffer from some limitations, since when their activation values are zero, the ReLUs cannot learn via gradient-based methods. To address this problem, the leaky ReLU (LReLU) activation function [51] with the following formula was used in this study:

$$F_{lrelu} = \begin{cases} x & x > 0 \\ 0.01x & otherwise \end{cases}$$

For feature reduction, the average pooling layer was used to summarize data at different feature map sub-regions. Suppose $C_Y$ is a collection of feature map values in the target pooling sub-region $Y$. Then, the average pooling $f_A$ is defined as follows according to [52]:

$$f_A = \frac{\sum C_Y}{|C_Y|}$$

The output of the last block was given to a dropout layer and finally a fully connected layer (i.e., Dense layers). The dropout layer solves the overfitting issue in CNNs by randomly setting a fraction of feature map values to 0, at each training iteration, which prevents them from relying on specific inputs [53]. Figure 1 shows the proposed CNN configuration. As the figure indicates, the input is different cost, medical and visit time series features described in Table 1. Medical cost, pharmacy cost and diagnosis code group $i$, procedure group $j$, drug group $k$ and visit group $m$ are some examples of these time series features. The proposed CNN consists of 128 filters in the first layer, 64 filters in the second layer and 32 filters in the third layer. Since the pooling size is two, the input size of each block is half the input size of the previous block. In the last step, different feature maps (representing different patterns) are combined to feed a fully connected layer.

In the evaluation step, we conducted five experiments to 1) find the best CNN architecture by performing hyper-parameter tuning; 2) compare the proposed CNN architecture against the common CNN architectures for pattern detection in images; 3) compare the proposed CNN pattern detection power against the most commonly used pattern detection methods for healthcare cost prediction; 4) evaluate the contribution of different predictors (i.e., cost, visit and medical features) on the proposed CNN performance; and 5) externally validate the proposed CNN. Each of these experiments is described in the following four subsections. Our data and evaluation setup are described in the last two subsections. An overview of the proposed method, experiments and evaluation is shown in Figure 1.

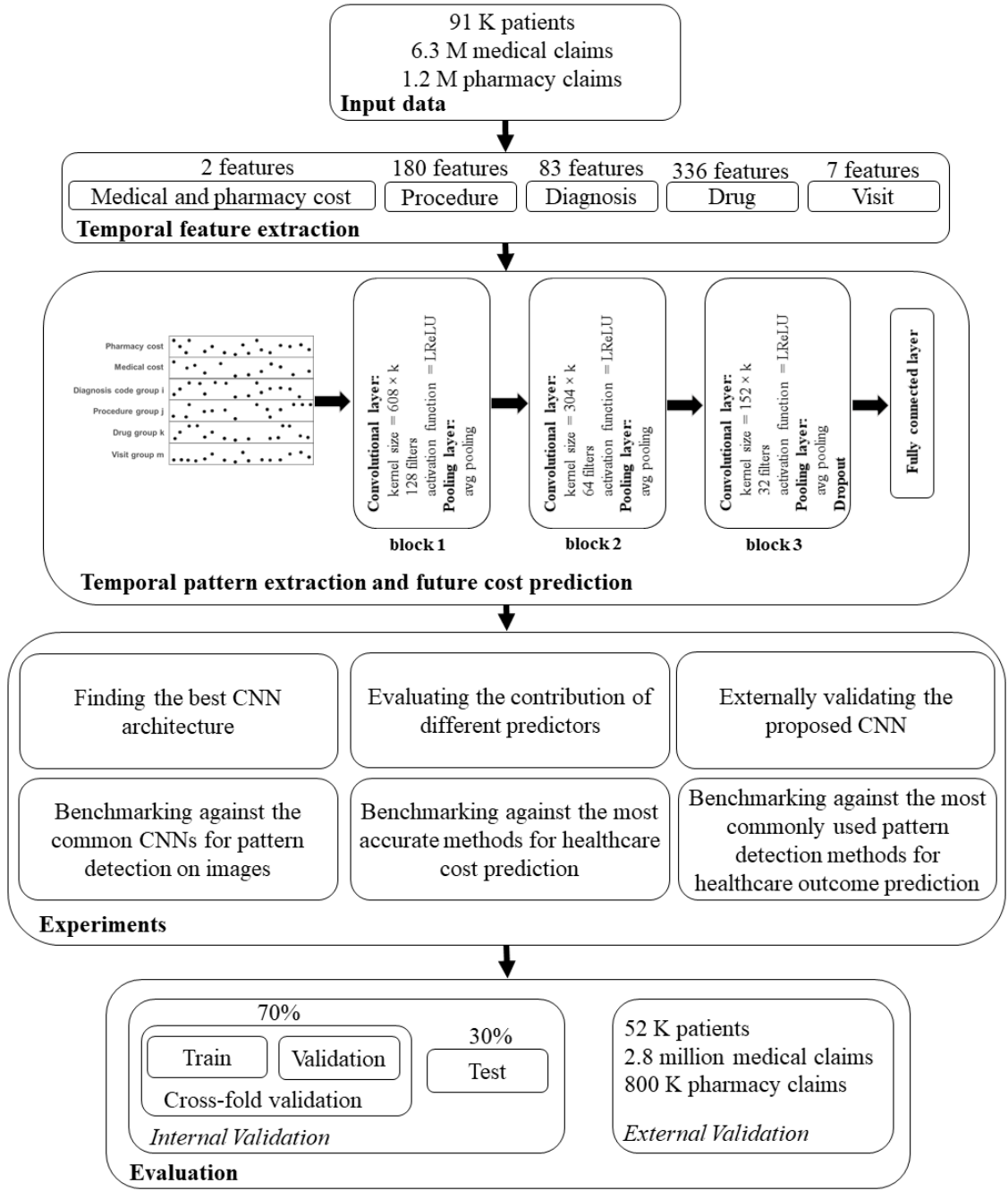

Figure 1. An overview of the proposed method, experiments and evaluation

**3.1 CNN hyper-parameter tuning**

Several hyper-parameters in CNNs should be tuned to find their best value. In this experiment, we reported the tuning results of the hyper-parameters that we found to be effective, including optimizing kernel size, number of filters, number of blocks, activation function and dropout rate. As mentioned, the kernel size in our proposed CNN took the form of *num_of_features* × *k*. Since *num_of_features* is fixed per block, we optimized *k* in this experiment. The number of combinations of potential settings for the various hyper-parameters, e.g., window size, number of blocks, *k*, type of activation function, drop-out rate and learning-rate, is too large for an exhaustive simultaneous search for the best combinations of hyper-parameter values. We therefore sequentially and iteratively tuned hyper-parameters. When selecting the appropriate value of a single hyper-parameter, the rest of the hyper-parameters were fixed. We tuned hyper-parameters in their order importance, e.g., *k* first. Once we obtained new results for the optimal values of some hyper-parameters (e.g., drop-out rate and activation function), we also retuned a previously-tuned hyper-parameter, e.g., *k*, by revising the fixed values of hyper-parameters to the new results in the *k*-retuning experiments. This process continued until the performance changes were negligible.

In all experiments, the following parameters were kept the same: each block consisted of a convolution layer; a LReLU layer; and a pooling layer, for which the convolution layer padding was the *"same"*, the stride was one, and for the pooling layer, the pooling size and its stride were two. Moreover, the Adam optimizer [54] was used to minimize mean squared error in the proposed CNN. All experiments were implemented in Python using the TensorFlow library.

## 3.2 Proposed CNN architecture vs. CNN architectures used for pattern detection in images

This experiment attempted to show that patients' multivariate time series that are framed as images are different from regular images, and therefore classical CNNs (proposed to detect patterns from regular images) are not the most effective methods to extract their patterns. As mentioned, AlexNet [35], VGG-16 [15] and ResNet-34 [39] are the commonly used CNN architectures for pattern detection in images. Therefore, we compared the performance of the proposed CNN architecture against these methods using transfer learning.

***Hypothesis 1:*** *The proposed CNN architecture is more accurate than the selected CNN architectures used for pattern detection in images.*

To implement AlexNet [35], VGG-16 [15] and ResNet-34 [39] CNN architectures with transfer learning, we re-trained the weights on our data, where the input was the patients' multivariate time series matrices rather than regular images (e.g., ImageNet data). More specifically, for each of the three CNNs, the architecture remained the same (e.g., same number of layers, same filters, same kernel size), but weights were re-trained on our data using a feed-forward approach to fix or re-train the weights in different layers with back propagation. Lastly, the final Dense layer was adjusted to have one neuron to predict cost rather than the image category. This approach is aligned with previous studies that used these methods as a baseline [14].

**3.3 Proposed CNN vs. spike pattern detection and symbolic pattern detection**

In the third experiment, our proposed CNN for healthcare cost prediction with temporal pattern detection was compared against two baselines: symbolic pattern detection as the most common method to implement temporal pattern detection and spike pattern detection as the most recent (state-of-art) method for cost prediction (i.e., our recent paper [7]) (see section 2.2). As mentioned, these methods are limited to a pre-defined set of pattern shapes that limit their performance. Because the proposed CNN can learn a variable number of patterns with various shapes, we hypothesize that the proposed CNN will outperform the selected benchmarks here.

***Hypothesis 2:*** *The proposed CNN method for temporal pattern detection is more accurate than pattern detection methods that are limited to a pre-defined set of pattern shapes.*

Aligned with Moskovitch and Shahar [31], we used the KLS framework to implement symbolic pattern detection. This framework includes three main components: temporal abstraction, time-interval mining, and TIRP-based and feature representation, where each component has its own settings. Aligned with Moskovitch and Shahar's suggestion, we used the following parameter settings after trying different settings in several evaluations: SAX for temporal discretization with four bins, KarmaLego with epsilon value of 0, and a minimal vertical threshold of 60% for three time-intervals mining. Three abstract relations (i.e., before, overlaps, and contains) proposed by the authors were used for temporal relations, and mean duration was used to represent TIRPs (without any feature selection). We used KarmaLego for feature learning (i.e., symbolic pattern detection), but for the numeric prediction (in the last stage), we used gradient boosting, which is the most successful cost prediction model in the literature [1,6,7].

In our most recent study [7], we showed that spike pattern detection outperforms the approach used in five prior research studies [1,3,5,17,18] on healthcare cost prediction identified from our literature review in [6]. The benchmark comparison of the proposed CNN against these studies are included in Table S1 of the online supplement.

Although our study focuses on predicting the actual numeric cost, to give a better sense of the importance of the temporal pattern detection methods on various cost buckets, we compared their performance in terms of classification measures as well. To achieve this, the final predicted cost (numeric) is mapped to one of the five cost buckets. This approach is aligned with the evaluation approach of previous cost prediction studies [3,7].

### 3.4 Contribution of predictors

In order to shed light on the category-level importance of various inputs to the proposed model, the fourth experiment attempted to show the effect of different predictors, including cost, visit and medical information on future cost prediction, by feeding different combinations of these predictors as inputs to the proposed CNN. This experiment can shed light on the effects of different types of features on cost prediction.

### 3.5 External validation

To externally validate the proposed CNN, we compared it to a cohort of patients from a separate insurance company. The validation was done by exporting the final CNN model extracted from the internal data set and applying it on the external data set. The same baseline models used for the internal validation (described above) were evaluated against the external data set with the same experimental setting.

### 3.6 Data

Our data set consisted of 1.2 million pharmacy claims and 6.3 million medical claims from approximately 91,000 distinct individuals covered by University of Utah Health Plans from October 2013 to October 2016. The study was approved by the University of Utah IRB (protocol #00094358). Available data included demographic information (e.g., age, gender), diagnosis and procedure codes, pharmacy dispense information, clinical visit information (e.g., place and date of service, provider information), and cost information (e.g., allowed, paid, and billed amount). Table 2 shows the demographic and insurance profile of the patients [7]. This data set is the same as the one used in our previous studies to benchmark state-of-art healthcare cost prediction methods [6,7].

Table 2 – Demographic and insurance profile of patients included in the internal analysis [7].

|  |  | **Number of members** | **% of members** |
|---|---|---|---|
| Age | 0-20 | 57,765 | 63% |
|  | 20-40 | 18,192 | 20% |
|  | 40-60 | 9,489 | 10% |
|  | 60-80 | 4,446 | 5% |
|  | 80-100 | 1,381 | 2% |
| Gender | Female | 50,931 | 56% |
|  | Male | 40,343 | 44% |
| Primary Insurance Provider | Yes | 74,311 | 81% |
|  | No | 16,963 | 19% |
| Insurance Type | Medicaid | 87,623 | 96% |
|  | Commercial | 3,651 | 4% |

Aligned with previous studies [3,7], we divided the data into two time periods: an observation period and a result period. The former time period, which was two years from October 2013 to September 2015, was used to predict individuals' cost in the one-year result period spanning October 2015 to October 2016. The transaction cost of a particular medical service or product (e.g., visit, prescription, or lab test) was measured in terms of the dollar amount the insurance company paid, which agrees with the approach in most of the related papers in the cost prediction literature [1,3,17], The target variable for medical cost prediction is then an aggregation of these costs that were incurred over the result period.

Since 80% of the overall cost of the population in the result period came from only 15% of the members in our data set, aligned with the literature on cost bucketing [3,7], the data set was partitioned into five cost buckets, with buckets 1 and 5 corresponding to the lowest and highest cost buckets, respectively. This partitioning was done so that the total dollar amount in each bucket was the same. Using cost bucketing, the performance of any proposed cost prediction method on low, moderate and high cost patients can be evaluated separately. Usually, predicting high cost patients is much more difficult than predicting low cost patients [1]. Also, it reduces the effects of outliers on the performance results.

For the external validation of our method, we used a data set consisting of around 800,000 pharmacy claims and 2.8 million medical claims from approximately 52,000 individuals covered by an insurance company in California from January 2015 to January 2018. Table 3 shows the demographic and insurance profile of these individuals. This data set provided information similar to that in our internal data set. January 2015 to December 2016 was used as the observation period, and January 2017 to January 2018 was used as the result period. Finally, the same cost bucketing method was employed on this data set to have the same total dollar amount in each bucket. All experiments were run by the insurance company using the model shared by the authors of this study.

Table 3 – Demographic and insurance profile of patients included in the external analysis.

|  |  | **Number of members** | **% of members** |
|---|---|---|---|
| Age | 0-20 | 21,366 | 41% |
|  | 20-40 | 15,113 | 29% |
|  | 40-60 | 9,902 | 19% |
|  | 60-80 | 3,648 | 7% |
|  | 80-100 | 2,085 | 4% |
| Gender | Female | 23,972 | 46% |
|  | Male | 28,142 | 54% |
| Primary Insurance Provider | Yes | 45,860 | 88% |
|  | No | 6,254 | 12% |
| Insurance Type | Medicaid | 13,549 | 26% |
|  | Commercial | 38,565 | 74% |

**3.7 Evaluation setup**

Cross validation was used on 70% of the data (randomly selected) for hyper-parameter tuning and selecting the optimal CNN architecture. The hyper-parameter tuning results on the validation data can be found in Tables S2 through S5 and Figure S1 in the online supplement. The remaining 30% of the data was used for the final evaluation and reporting of results. The numeric cost prediction performance was measured according to the average of mean absolute percentage error (MAPE) across 20-fold cross validation, which is the most common relative error measure for cost prediction used in the literature [6]:

$$\frac{1}{N} \sum_{i=1}^{N} \left| \frac{Actual_i - Predicted_i}{Actual_i} \right|$$

where *N* is the number of instances (i.e., patients). To avoid division by zero, we added one dollar to each patient's actual cost in the result period. We could not report error measures that use absolute values (e.g., mean absolute error) due to the sensitive nature of the data.

MAPE was evaluated across the 20-fold validation using pair-wise t-tests, which results in a large number of comparisons. Therefore, a Bonferroni correction [55] was performed as a post hoc test, where only p-values less than 0.01 were considered to be statistically significant at an alpha = 0.05. This statistical approach was aligned with the method recommended by Demsar [56].

For the third experiment, in which we compared the proposed CNN pattern detection power against the most commonly used pattern detection methods for healthcare cost prediction (i.e., literature baselines), the classification performance was assessed in terms of accuracy, recall, precision and a domain knowledge based measure called penalty error [3] in addition to MAPE. The penalty error penalizes the misclassification of various cost-buckets differently. For instance, it penalizes models for underestimating high cost members or overestimating low cost members. We adopted the same penalty table used in previous cost prediction studies [3,7] (see Table S6 in the online supplement).

# 4. RESULTS

## 4.1 CNN hyper-parameter tuning

As shown in Table 4, the MAPE performance of the CNN with three blocks with 128, 64 and 32 filters was significantly better than those of the CNNs with all other combinations of the numbers of filters in one to five blocks (p<0.01 in all comparisons).

Table 4 - Effect of the number of filters and the number of blocks in the CNNs on the MAPE performance of cost prediction over the five cost buckets.

| Number of blocks | Numbers of filters | all | 1 | 2 | 3 | 4 | 5 |
|---|---|---|---|---|---|---|---|
| 1 | 16 | 6.28 | 6.24 | 6.41 | 6.53 | 6.67 | 7.74 |
| 1 | 32 | 5.52 | 5.47 | 5.59 | 5.87 | 6.01 | 7.57 |
| 1 | 64 | 4.08 | 4.02 | 4.34 | 4.47 | 4.74 | 5.04 |
| 1 | 128 | 4.55 | 4.45 | 4.63 | 5.58 | 5.74 | 6.87 |
| 1 | 256 | 4.74 | 4.64 | 4.97 | 5.47 | 6.32 | 6.45 |
| 2 | 16, 8 | 4.33 | 4.19 | 4.94 | 5.18 | 5.69 | 6.24 |
| 2 | 32, 16 | 2.75 | 2.67 | 3.01 | 3.12 | 4.27 | 4.63 |
| 2 | 64, 32 | 2.61 | 2.52 | 2.94 | 3.08 | 4.19 | 4.52 |
| 2 | 128, 64 | 2.64 | 2.54 | 2.97 | 3.14 | 4.24 | 4.53 |
| 2 | 256, 128 | 4.76 | 4.69 | 4.81 | 5.64 | 5.96 | 6.02 |
| 3 | 32, 16, 8 | 3.11 | 3.08 | 2.94 | 3.76 | 3.97 | 4.17 |
| 3 | 64, 32, 16 | 2.23 | 2.13 | 2.67 | 2.97 | 3.09 | 3.67 |
| **3** | **128, 64, 32** | **1.67** | **1.58** | **2.03** | **2.35** | **2.72** | **2.98** |
| 3 | 256, 128, 64 | 3.37 | 3.24 | 3.96 | 4.19 | 4.78 | 5.29 |
| 3 | 512, 256, 128 | 4.79 | 4.66 | 5.17 | 5.94 | 6.03 | 6.97 |
| 4 | 64, 32, 16, 8 | 3.40 | 3.28 | 3.76 | 4.19 | 5.24 | 6.03 |
| 4 | 128, 64, 32, 16 | 3.17 | 3.04 | 3.54 | 4.08 | 4.96 | 5.62 |
| 4 | 256, 128, 64, 32 | 4.35 | 4.21 | 4.76 | 5.33 | 6.29 | 6.94 |
| 4 | 512, 256, 128, 64 | 4.85 | 4.77 | 4.81 | 5.55 | 6.76 | 7.29 |
| 5 | 128, 64, 32, 16, 8 | 3.56 | 3.41 | 3.77 | 5.14 | 5.69 | 5.92 |
| 5 | 256, 128, 64, 32, 16 | 4.82 | 4.71 | 4.99 | 5.92 | 6.19 | 6.85 |
| 5 | 512, 256, 128, 64, 32 | 5.21 | 5.07 | 5.63 | 6.27 | 6.79 | 7.51 |

As shown in Table 5, the MAPE performance of $k = 3$ time windows outperformed k = 1, 6 and 9 time windows (1.67 versus 4.62, 2.20 and 3.5; p<0.01).

Table 5 - MAPE of different number of time windows, $k$, used in CNNs for cost prediction over the five cost buckets.

| K | all | 1 | 2 | 3 | 4 | 5 |
|---|-----|-----|-----|-----|-----|-----|
| 1 | 4.62 | 4.54 | 4.84 | 5.23 | 5.99 | 6.04 |
| **3** | **1.67** | **1.58** | **2.03** | **2.35** | **2.72** | **2.98** |
| 6 | 2.20 | 2.11 | 2.55 | 2.72 | 3.07 | 3.61 |
| 9 | 3.50 | 3.26 | 4.66 | 5.01 | 5.48 | 5.69 |

As shown in Table 6, the MAPE performance of LReLU activation function outperformed ReLU, Sigmoid and Tanh activation functions (1.67 versus 2.21, 3.67 and 5.44; p<0.01).

Table 6 - MAPE of different activation functions used in CNNs for cost prediction over the five cost buckets.

| Activation Function | all | 1 | 2 | 3 | 4 | 5 |
|---|-----|-----|-----|-----|-----|-----|
| **LReLU** | **1.67** | **1.58** | **2.03** | **2.35** | **2.72** | **2.98** |
| ReLU | 2.21 | 2.06 | 2.78 | 3.09 | 3.84 | 4.16 |
| Sigmoid | 3.67 | 3.47 | 4.53 | 4.99 | 5.64 | 6.13 |
| Tanh | 5.44 | 5.21 | 6.47 | 6.99 | 7.55 | 8.21 |

As shown in Table 7, the Dropout rate of 0.5 had the highest MAPE performance compared to other rates (p<0.01 in all comparisons).

Table 7 - MAPE of different dropout rates used in CNNs for cost prediction over the five cost buckets.

| Dropout Rate | all | 1 | 2 | 3 | 4 | 5 |
|---|---|---|---|---|---|---|
| 0.75 | 3.95 | 3.79 | 4.61 | 5.09 | 5.74 | 6.11 |
| **0.5** | **1.67** | **1.58** | **2.03** | **2.35** | **2.72** | **2.98** |
| 0.25 | 2.73 | 2.59 | 3.27 | 3.83 | 4.18 | 4.59 |
| 0.15 | 3.72 | 3.64 | 3.92 | 4.19 | 4.89 | 5.61 |
| 0 (no dropout) | 5.29 | 5.08 | 6.11 | 6.84 | 7.23 | 8.64 |

**4.2 Proposed CNN architecture vs. CNN architectures used for pattern detection in images**

As shown in Table 8, the MAPE performance of the proposed CNN was significantly better than that of ResNet-34 [39], AlexNet [35] or VGG-16 [15] (1.67 versus 4.59, 4.85 and 5.06; p<0.01).

Table 8 - MAPE of the proposed CNN architecture versus commonly used CNN architectures for pattern detection in images over the five cost buckets.

| Method | all | 1 | 2 | 3 | 4 | 5 |
|---|---|---|---|---|---|---|
| Proposed CNN | **1.67** | **1.58** | **2.03** | **2.35** | **2.72** | **2.98** |
| ResNet-34 [39] | 4.59 | 4.42 | 5.48 | 5.56 | 5.96 | 6.29 |
| AlexNet [35] | 4.85 | 4.67 | 5.76 | 6.01 | 6.36 | 6.47 |
| VGG-16 [15] | 5.06 | 4.89 | 5.84 | 6.23 | 6.49 | 6.81 |

**4.3 Proposed CNN versus spike pattern detection and symbolic pattern detection**

As shown in Table 9, the MAPE performance of the proposed CNN outperformed spike pattern detection versus symbolic pattern detection for cost prediction (1.67 versus 2.02 and 3.66; p<0.01).

The proposed CNN was significantly more accurate than spike pattern detection and symbolic pattern detection for cost prediction (94.53 versus 90.03 and 87.02; p<0.01). Similar significant performance differences were found in terms of precision and recall (Table 10) as well as penalty error (Table 11).

Table 9 - MAPE of the proposed CNN versus spike pattern detection and symbolic pattern detection for cost prediction over the five cost buckets.

| Method | all | 1 | 2 | 3 | 4 | 5 |
|---|---|---|---|---|---|---|
| Proposed CNN | **1.67** | **1.58** | **2.03** | **2.35** | **2.72** | **2.98** |
| Spike pattern detection | 2.02 | 1.94 | 2.25 | 2.53 | 3.07 | 4.21 |
| Symbolic pattern detection | 3.66 | 3.47 | 4.59 | 4.87 | 5.01 | 5.57 |

Table 10 – Accuracy, precision and recall of the proposed CNN versus spike pattern detection and symbolic pattern detection for cost prediction over the five cost buckets.

| Method | Accuracy | Recall | | | | | Precision | | | | |
|---|---|---|---|---|---|---|---|---|---|---|---|
| | | 1 | 2 | 3 | 4 | 5 | 1 | 2 | 3 | 4 | 5 |
| Proposed CNN | 94.53 | 96.8 | 86.4 | 76.7 | 68.1 | 64.2 | 96.7 | 87.2 | 76.6 | 68.8 | 65.2 |
| Spike pattern detection | 90.03 | 94.9 | 66.8 | 57.5 | 50.4 | 48.2 | 95.2 | 67.1 | 58.7 | 51.3 | 49.7 |
| Symbolic pattern detection | 87.02 | 92.9 | 57.7 | 50.2 | 38.2 | 34.3 | 93.1 | 58.3 | 48.9 | 40.3 | 25.5 |

Table 11 – Penalty error of the proposed CNN versus spike pattern detection and symbolic pattern detection for cost prediction over the five cost buckets.

| Method | all | 1 | 2 | 3 | 4 | 5 |
|---|---|---|---|---|---|---|
| Proposed CNN | 0.12 | 0.09 | 0.24 | 0.29 | 0.37 | 0.42 |
| Spike pattern detection | 0.23 | 0.17 | 0.54 | 0.59 | 0.66 | 0.70 |
| Symbolic pattern detection | 0.59 | 0.52 | 0.91 | 0.95 | 1.14 | 1.36 |

**4.4 Contribution of predictors**

As shown in Figure 2, all of the predictors combined had a more significant contribution to the CNN's cost prediction performance than any partial combinations of the predictors by category. Cost predictors were the most important features, followed by medical and visit predictors (MAPE of 2.04 versus 1.98 and 1.91; p<0.01).

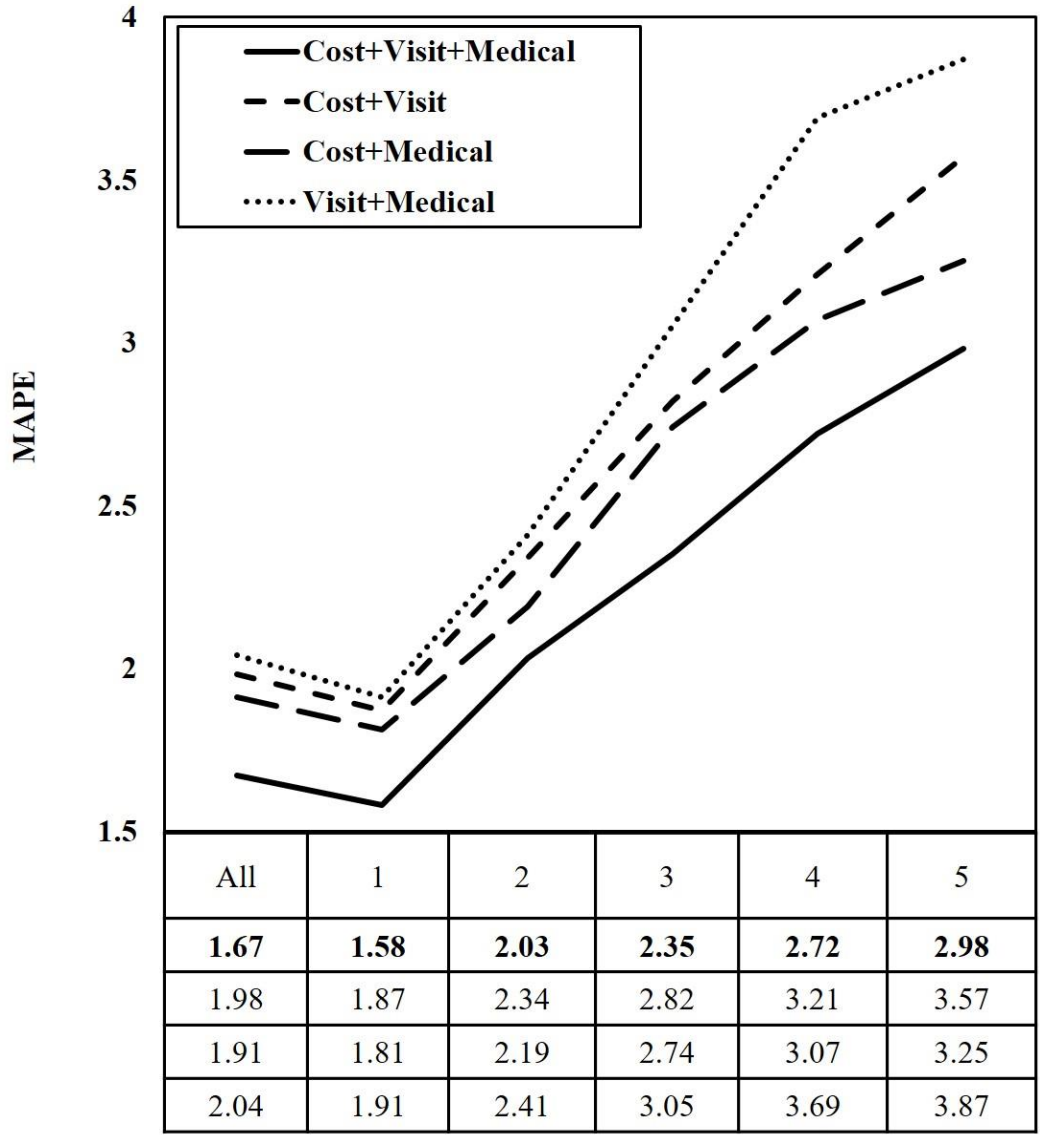

| | All | 1 | 2 | 3 | 4 | 5 |
|---|---|---|---|---|---|---|
| **Cost+Visit+Medical** | **1.67** | **1.58** | **2.03** | **2.35** | **2.72** | **2.98** |
| Cost+Visit | 1.98 | 1.87 | 2.34 | 2.82 | 3.21 | 3.57 |
| Cost+Medical | 1.91 | 1.81 | 2.19 | 2.74 | 3.07 | 3.25 |
| Visit+Medical | 2.04 | 1.91 | 2.41 | 3.05 | 3.69 | 3.87 |

Figure 2 - MAPE of the proposed CNN over different types of predictors for cost prediction over the five cost buckets.

**4.5 External validation**

As shown in Table 12, the MAPE performance of the proposed CNN outperformed spike pattern detection, symbolic pattern detection, ResNet-34 [39], AlexNet [35], and VGG-16 [15] on the external data set (1.82 versus 2.18, 3.88, 4.86, 5.09 and 5.30; p<0.01). Table S7 shows both internal and external validation results.

Table 12 - MAPE of the proposed CNN architecture versus all baselines over the five cost buckets on the external data set.

| Method | all | 1 | 2 | 3 | 4 | 5 |
| --- | --- | --- | --- | --- | --- | --- |
| **Proposed CNN** | **1.82** | **1.68** | **2.13** | **2.55** | **2.89** | **3.17** |
| Spike pattern detection | 2.18 | 2.06 | 2.37 | 2.68 | 3.23 | 4.38 |
| Symbolic pattern detection | 3.88 | 3.61 | 4.69 | 5.03 | 5.16 | 5.73 |
| ResNet-34 [39] | 4.86 | 4.62 | 5.67 | 5.77 | 6.19 | 6.53 |
| AlexNet [35] | 5.09 | 4.82 | 5.93 | 6.26 | 6.59 | 6.72 |
| VGG-16 [15] | 5.30 | 5.05 | 6.04 | 6.44 | 6.74 | 7.03 |

# 5. DISCUSSION

In this study, we investigated a method for leveraging patients' temporal data using deep learning methods to predict healthcare costs. Predicting the insured patients' costs using accurate prediction methods is important for various groups of people. First, knowing in advance that they are at high risk of an increase in expenditures for the next year, patients could choose insurance plans with expanded coverage. Second, insurance companies can more effectively control costs if they can more accurately predict and manage patients who will incur the greatest costs. Third, if healthcare service providers (e.g., doctors and hospitals) operate under at-risk payment models and know which of their patients are going to incur high costs, they could also take measures to reduce costs, such as by arranging more frequent check-ups or better coordinating care.

This study makes three main contributions. First, we showed how to benefit from deep learning methods (specifically CNN) for leveraging latent 2D temporal patterns from multivariate time series patient data. Since CNN is usually used for image data, understanding how it can be adjusted for temporal patient data fills the important unmet need to apply deep learning to patient cost prediction. Second, this study explored and validated the effectiveness of CNN's feature learning capability for pattern detection for patient cost prediction. We showed CNN can learn patterns that have higher predictive power than the patterns previously identified by the patient cost prediction literature. Third, we demonstrated the effectiveness of all medical, visit and cost features for cost prediction. More specifically, whereas the previous literature did not find medical features to be effective predictors compared to cost features, this paper showed their effectiveness. This study also suggests CNN's feature learning capacity could be useful for other healthcare contexts that involve patient outcome prediction using multivariate time series data.

We conducted one set of experiments for hyper-parameter tuning of CNNs to find the best architecture, one experiment to show the advantage of the proposed CNN architecture over some of the award-winning CNNs for image classification, one experiment to show the advantage of CNN over the state-of-art temporal pattern detection methods for cost prediction, one experiment to show the contribution of each type of predictor (i.e., cost, medical and visit) for cost prediction, and one experiment to externally validate the proposed CNN.

The first experiment showed that having three blocks of convolution and pooling layers with LReLU activation using 128, 64 and 32 filters demonstrates the best performance. The MAPE performance degraded when the CNN architecture was too shallow or deep. In shallow CNNs, increasing the number of filters benefited prediction accuracy. The number of filters corresponds to the variety of 2D multivariate temporal patterns each CNN block produces. Hence, the performance of patient cost prediction deteriorated when pattern learning became ineffective with too few or too many filters.

The depth of a CNN helps enhance the hierarchical composition of the hidden 2D patterns from low-level patterns to advanced patterns. However, if the training sample is not sufficiently large, a deep CNN architecture cannot lead to advanced feature learning effectively, and hence it does not benefit the prediction performance. Moreover, we evaluated the effect of the number of time windows in the kernels by changing it from 1 to 9 (other settings were the same as in the previous experiment). The results showed the advantage of using three for the number of time windows in the kernels. One insight is that three-month periods (e.g., last three months' cost) have a significant effect on future costs, as reported in previous studies [3]. One month is too short, whereas six and nine months negatively impacted the performance due to higher levels of temporal pattern abstraction. Also, the results confirmed the advantage of using Leaky ReLU over other activation functions due to gradient-based learning. Finally, this experiment showed that a dropout layer with proper adjustment is needed to avoid overfitting the proposed CNN to the training data. These results indicated the importance and benefit of exploring the decisions of hyper-parameter choices in configuring a CNN architecture for learning latent 2D patterns from multivariate time series input data.

In the second experiment, the proposed CNN architecture outperformed ResNet [39], AlexNet [35] and VGGNet [15], as the most common architectures for pattern detection in images. This result shows that multivariate time series data shaped as images are different from regular images, and hence they need a unique kind of architecture. In other words, small kernels (e.g., 3x3) and advanced architectural configurations designed for conventional image data tend not to help CNNs learn effective 2D patterns from multivariate time series data. Since the order of the features along the variable dimension does not have the same meaning as the order of pixels in a conventional image, the choice of a kernel of *number_of_features* $\times$ *3* time windows in the proposed CNN relaxed the narrow focus of the adjacent features resulting from the 2D kernels of the small sizes used in ResNet [39], AlexNet [35] and VGGNet [15]. The experiment results suggest that a larger sample size of patients' multivariate time series data would be necessary before an advanced CNN architecture, like the award-winning CNNs, has the potential to predict more accurately than the proposed CNN. This experiment confirmed hypothesis H1.

In the third experiment, we compared the proposed CNN with the KarmaLego method [31], which is the most common approach for temporal pattern detection in healthcare data, and the spike pattern detection method [7], which is detailed in the most recent study on cost prediction and has the best performance compared to previous cost prediction studies. The proposed CNN outperformed both of these methods. We believe the main reason for this finding is that these two baseline methods (and all other similar methods) attempt to detect a pre-defined set of pattern shapes in healthcare data, whereas CNN can learn a variety of pattern shapes and even extract patterns hierarchically. Hence, the results confirm hypothesis H2 and demonstrate the benefits of automatically learning latent 2D temporal patterns for patient cost prediction over models with manually-crafted temporal patterns.

In the fourth experiment, we attempted to assess the effect of different types of predictors (i.e., cost, medical and visit) on the performance of the proposed CNN. All three types of predictors made significant contributions to the final performance of the proposed CNN for cost prediction. Although the effectiveness of cost and visit predictors had been found in previous studies, our study also identified the effectiveness of medical predictors. The implication is that the proposed CNN was able to find some patterns from medical predictors that could not be uncovered by the previous methods. Also, while in the past, studies removing cost predictors had a large negative effect on performance, this was not the case in this study. This shows that finding proper patterns from medical and visit predictors can replace cost predictors to some extent. The results in Figure 2 also suggest that, even with a reduced variety of predictors and lower data acquisition efforts for training the cost predictive model, the proposed CNN still has more accurate predictions than other benchmarks that have used pre-defined temporal patterns extracted from the full set of predictors.

In the last experiment, the proposed CNN was externally validated against a secondary data set, and the performance was found to be aligned with that for the primary data set used for internal validation. Also, similar superiority of the proposed model against the same baselines (used for the internal validation) further validated its robustness against the external data set.

The main focus of this paper was on enhancing the accuracy of the cost prediction methods with state-of-art machine learning methods, which has several important practical applications, as explained before. The explainability of such methods could enhance their usability and acceptance in the clinical setting. Although this study showed the category-level importance of the input features to the proposed CNN model, a limitation of this study is the lack of individual-level explainability of the input features' contributions to the proposed model. Whereas CNN models with categorical target variables can be explained to some extent by using visualization techniques, CNNs for numeric prediction are more limited. Moreover, another potential future direction is to study the effect of other deep learning models such as recurrent neural networks (RNNs) for cost prediction, since they have also been shown to be effective for time series forecasting. Such future research on RNNs for cost prediction requires investigation of how patients' cost in the past can predict their future costs (in a single time series), as well as how a combination of multiple individual time series of other input can enhance the effectiveness of patient cost forecasting.

## 6. CONCLUSION

In this paper, we attempted to improve the performance of healthcare cost prediction methods by leveraging the feature learning power of convolutional neural networks for temporal pattern detection. The proposed CNN was able to improve the performance of the best methods in the literature in temporal pattern detection by automatically learning latent 2D temporal patterns. Also, our study confirmed that proper extraction of temporal patterns can enable cost, visit and medical data to be effective predictors of healthcare cost. The proposed CNN in this study could be extended to other healthcare domains to predict other health outcomes of interest by leveraging multivariate time series of patient data.

**ACKNOWLEDGEMENTS**

The author would like to thank their collaborators in the University of Utah Health Plans, Mr. Travis Ault and Ms. Josette Dorius, for their help with data set acquisition, preparationh and interpretation, as well as for their guidance on the clinical implications of this work.